\documentclass[runningheads]{llncs}
\usepackage[T1]{fontenc}
\usepackage{graphicx}
\usepackage{booktabs}
\usepackage[misc]{ifsym}
\newcommand{\corr}{(\Letter)}
\usepackage{mwe}
\usepackage{amsmath}
\usepackage{amsfonts}
\usepackage{subfigure}
\usepackage{hyperref}
\usepackage{multirow}

\begin{document}

\title{Colliding with Adversaries at ECML-PKDD 2025 
\\ Adversarial Attack Competition 
 \\ 1st Prize Solution
}

\titlerunning{ Colliding with Adversaries  -  Adversarial Attack 1st Prize Solution}


\author{Dimitris Stefanopoulos\inst{1} \corr \and Andreas Voskou\inst{2} }

\institute{Aristotle University of Thessaloniki \email{dstefanop@math.auth.gr}  \and  Cyprus University of Technology, Limassol, Cyprus  \email{ai.voskou@edu.cut.ac.cy} } 

\maketitle 

\begin{abstract}
This report presents the winning solution for Task 1 of Colliding with Adversaries: A Challenge on Robust Learning in High Energy Physics Discovery at ECML-PKDD 2025. The task required designing an adversarial attack against a provided classification model that maximizes misclassification while minimizing perturbations. Our approach employs a multi-round gradient-based strategy that leverages the differentiable structure of the model, augmented with random initialization and sample-mixing techniques to enhance effectiveness. The resulting attack achieved the best results in perturbation size and fooling success rate, securing first place in the competition.
\end{abstract}



\section{Introduction}

This report describes the solution of team APTH for Task 1 of the Colliding with Adversaries: A Challenge on Robust Learning in High Energy Physics Discovery challenge of ECML-PKDD 2025. The task focuses on performing adversarial attacks against machine learning models trained on experimental high-energy physics data in tabular format \cite{saala2025introduction}. Specifically, it concerns a binary classification problem between two different simulated processes: (i) two top-Jets (TTJets) and (ii) two W-Boson-jets. The goal is to generate adversarial examples that force misclassifications by a given model (based on TopoDNN \cite{kasieczka2019machine},\cite{saala2025enforcing}), while keeping perturbations as small as possible.

Given that both the model and the evaluation dataset were provided, we adopted a white-box, model-aware attack methodology. Since TopoDNN is a differentiable neural network, allowing for gradient-based optimization, our approach centered on a gradient-descent-based attack.  The code for the proposed methodology can be found in the  \href{https://github.com/jmstf94/Colliding-with-Adversaries}{Colliding with Adversaries GitHub repository}.

\section{Task Definition}

Let the dataset be denoted as $D = \{X, Y\}$, where $X \in \mathbb{R}^{5000 \times 87}$ is the input matrix and $Y \in \{0,1\}^{5000}$ the corresponding binary labels. Each pair $(x_i, y_i)$ corresponds to a data point and its label for $i \in [1,5000]$. The model under attack is denoted as $F$, representing the TopoDNN-like network with the provided pretrained weights.

The objective is to find, for each input $x_i$, a perturbed input $x'_i = x_i + \delta x_i$ such that the model's prediction changes: $F(x_i) \neq F(x'_i)$, while keeping the perturbation $||\delta x_i||_1$ as small as possible.

The evaluation metric for the competition is defined as:

\begin{equation}
S = \text{FR} \cdot e^{-20 D}
\end{equation}
where $\text{FR}$ (Fooling Ratio) is the proportion of originally correctly classified samples that become misclassified after the attack, and $D = \frac{1}{N_f} \sum_{i=1}^{5000} ||\delta x_i||_1$ is the average $L_1$ distance between original and adversarial examples for the successfully reversed (fooled) cases. The total number of reversed predictions is denoted as $N_f$. The task therefore requires jointly (i) reversing the model's prediction and (ii) minimizing the input perturbation. This leads to a mixed optimization objective that balances adversarial effectiveness (via cross-entropy loss or similar) and minimization of the perturbation (via $L_1$ norm).

\section{Core Methodology}

Based on the task formulation and taking into account that:
\begin{enumerate}
\item The fooling ratio is binary per sample (1 if prediction flips, 0 otherwise),
\item The provided TopoDNN model ($F$) has 100\% accuracy on the clean dataset,
\item We only care about minimizing $ || \delta x_i ||_1$ for successful attacks (i.e., when the prediction flips), since the rest can be considered as having zero contribution.
\end{enumerate}
we designed the following objective function:
\begin{equation}
L =
\begin{cases} 
L_{fool} & \text{if } F(x_i + \delta x_i)  =  F(x_i) \\
L_{reduce} & \text{if } F(x_i + \delta x_i) \neq F(x_i)
\end{cases}
\end{equation}
where:
\begin{equation}
L_{\text{fool}} = - \text{BCE}(F(x_i + \delta x_i), y_i)
\end{equation}
\begin{equation}
L_{\text{reduce}} =||\delta x_i||_1
\end{equation}

Here, $L_{\text{fool}}$ encourages flipping the model's prediction using the negative binary cross-entropy as a minimization objective, but only if a prediction flip has not yet been achieved. This corresponds to targeting a prediction confidence just over 50\%, sufficient to change the prediction without requiring excessive perturbation. $L_{\text{reduce}}$ corresponds to direct $L_1$ distance minimization of $||\delta x||_1$, encouraging small perturbations; this term applies only when the prediction flip has already been achieved. This formulation prioritizes reversing the prediction and avoids unnecessary variance from optimizing both terms simultaneously.

After optimizing the main objective, we also perform a follow-up pure $L_{fool}$ optimization for a small number of steps to ensure that predictions have been successfully reversed and to avoid marginally non-reversed cases.

The objective minimization was performed using vanilla gradient descent. While more sophisticated approaches such as Adam could be faster or more effective, we were able to achieve the desired results easily, and GD proved sufficient.

\section{Overall Approach}

While the previously described process was the core component of our methodology, a few auxiliary strategies were employed to further improve the performance of our adversarial attacks.

Instead of running the optimization only once based on the original test set as a starting point, we performed it multiple times with several variations. Specifically, we executed a long sequence of optimization rounds, 150 in total, with 20 parallel runs in each. The optimization step size for each run was set randomly using a changing uniform distribution with decreasing maximum and minimum values:
\begin{align}
step_n^j \sim U(min_n,max_n) \\
max_n =  2 ^{(9 -\frac{n}{5})}, \;\; min_n = \frac{max_n}{10}
\end{align}
where $n \in [1,150]$ is the optimization round, and $j \in [1,20]$ denotes each of the 20 parallel runs. The step size for the follow-up optimization is always set to $step_{fu} = 0.01$. The main runs are performed using 2500 total steps, while the follow-up runs use 250 steps.

Furthermore, instead of always using the original data as the initial point, each of the 20 parallel runs for each optimization round used a different approach. Specifically, the following strategies were employed:

\begin{itemize}
\item For $j = 1$, we use the original data $X$.
\item For $j = 2$, we use the best state up to the current round, $X_{best}$.
\end{itemize}

For the remaining cases, we define and use a row-wise random mixture process formulated as:
\begin{equation}
f_{rm}(x_a,x_b) := x_a \cdot w + x_b \cdot (1-w), \;\; w \sim U(0,1)
\label{eq:mix}
\end{equation}
where $x_a, x_b \in \mathbb{R}^{87}$ and $w \in \mathbb{R}$. This method is equivalent to randomly selecting a  point on the line connecting $x_a$ and $x_b$ in the feature space. This technique is applied as follows:

\begin{itemize}
\item For $j = 3$ to $j = 10$, we mix the original data $X$ with $X_{best}$, generating 8 different random mixtures: $x_i^j = f_{rm}(x_i, x_{best_i})$.
\item For $j = 11$ to $j = 20$, we again use the mixture process in equation \eqref{eq:mix}, but with different components. Assuming that for a specific point $x_a$, the correct label is 1, we randomly select an example $x_b$ from the original dataset $X$ with the opposite label (in this case, 0), and randomly mix them. Technically, we leverage the fact that we know a direction where the prediction will eventually reverse and move the initial point toward it. This approach proved effective in the hardest-to-reverse cases, where local gradients do not provide reliable guidance for prediction reversing.
\end{itemize}

Finally, we note that at the end of each round, the initial conditions have a 50\% chance to be updated based on the procedure above, and a 50\% chance to continue from the previous state. After each round, the best result for each row (with a prediction flip and minimum $||\delta x_i||_1$), among the 20 runs or the previous best, is selected to form $X_{best}$. The final $X_{best}$ forms the  solution for the task.

\section{Evaluation and Discussion}

\begin{table}[h]
\caption{Results Comparison}
\centering
\renewcommand{\arraystretch}{1.1}
\begin{tabular}{lll}
\toprule
& \textbf{Method} & \textbf{Score}   \\
\midrule
\multirow{2}{*}{ Competition}& Third Place      & 0.91      \\
 &Second Place            & 0.94   \\
\midrule
 \multirow{3}{*}{ Ours} &Core   & 0.934    \\
 &Single Round    &      0.956      \\
 &Full Method      &    \textbf{0.976 }       \\
\bottomrule
\label{tab:results}
\end{tabular}
\end{table}

We evaluate our approach on the provided test set $X_{\text{test}}$, with results summarized in Table~\ref{tab:results}. For reference, we also report the second- and third-place solutions from the challenge.  

The "Core" variant corresponds to a baseline gradient-based optimization attack, applied for 50,000 steps with a learning rate schedule starting at 0.01 and increasing up to 100 every 10,000 steps. This yields a score of $S = 0.934$, slightly below the second-place result.  

\begin{figure}
\centering
\includegraphics[width=0.93\linewidth]{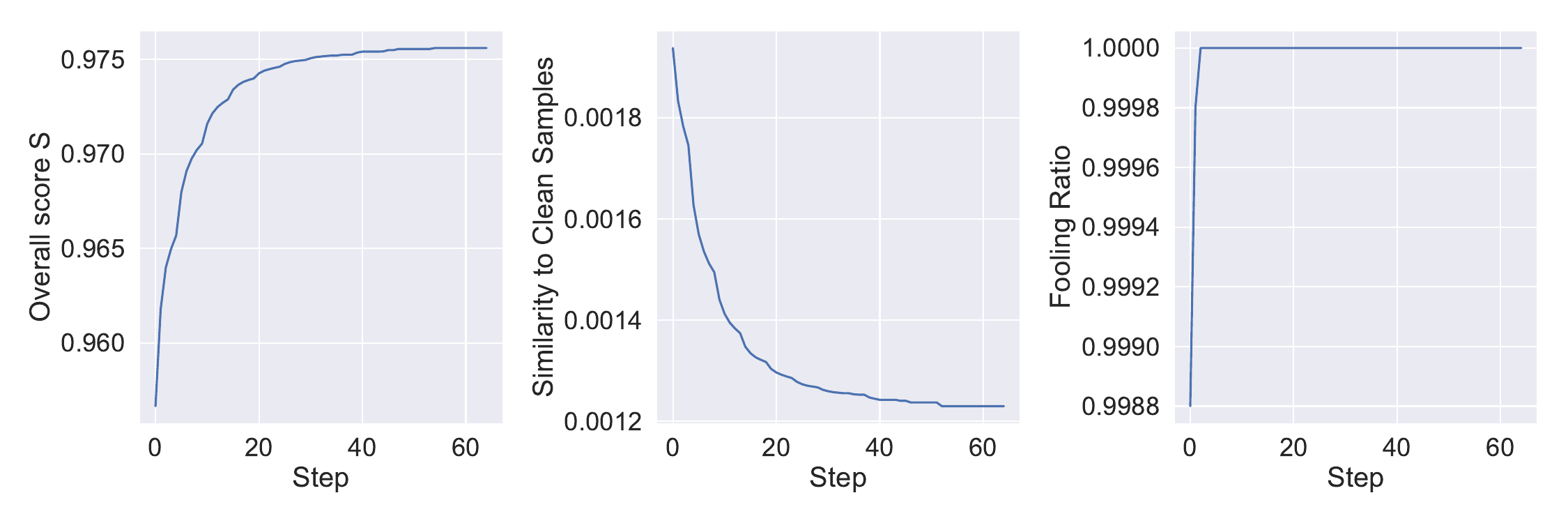}
\caption{Fooling metrics evolution per step.}
\label{fig:evo}
\end{figure}

Applying a single round of our full method improves performance to $S = 0.956$, already surpassing all competing solutions. Figure~\ref{fig:evo} further illustrates the progression of key attack metrics—overall score $S$, Fooling Ratio, and similarity to clean samples—across the first 60 rounds. The Fooling Ratio reaches 100\% within just a few rounds, while the distance from clean samples decreases smoothly from 0.002, stabilizing at about 0.0012 after 50 rounds. Ultimately, our full method achieves $S = 0.9756$, outperforming the second-best solution by approximately 0.04.

While our methodology has proven to be particularly effective for the specific task, it is not without limitations or room for improvement. First, achieving the full score requires multiple rounds of optimization, which may take several hours to complete. Second, since we perform 20 parallel runs per session, the approach necessitates access to GPU hardware with relatively high memory capacity. Finally, the proposed technique is a gradient-based white-box attack tailored to exploit vulnerabilities of the specific model under study; unlike model-agnostic shuffle-based attacks, it cannot be guaranteed to generalize well to other models, nor is it applicable to systems that do not rely on deep learning.

In addition, regarding potential improvements, our methodology relies on simple vanilla gradient descent without the use of momentum or adaptive terms. The use of more advanced optimization algorithms such as Adam represents an evident opportunity for enhancement. Moreover, our technique of mixing with known entries could be made more sophisticated or targeted, which may further improve the attack’s effectiveness and resulting scores.
\vspace{-1mm}

\section{Conclusion}

Using a gradient descent optimization method and further improving it through random yet targeted alterations of the initial starting points, we managed to achieve a particularly high score of $S = 0.98$ and secured first place in the competition, with a solid margin from the second-best solution.

\bibliographystyle{splncs04}
\bibliography{paper}

\end{document}